\documentclass[conference]{IEEEtran}
\usepackage{graphicx}
\usepackage[export]{adjustbox}
\usepackage{subfigure}
\usepackage{float}
\usepackage[T1]{fontenc}
\usepackage{tikz}
\usetikzlibrary{arrows,calc,positioning}
\begin{document}
\title{Automated image segmentation for detecting cell spreading for metastasizing assessments of cancer development}

\author{\IEEEauthorblockN{Sholpan~Kauanova, Ivan~Vorobjev, and Alex Pappachen James}
\IEEEauthorblockA{
School of Engineering, Nazarbayev University, Astana\\
www.biomicrosystems.info/alex\\
Email: skashikova@nu.edu.kz; apj@ieee.org}}

\maketitle
\begin{abstract}
The automated segmentation of cells in microscopic images is an open research problem that has important implications for studies of the developmental and cancer processes based on in vitro models. In this paper, we present the approach for segmentation of the DIC images of cultured cells using G-neighbor smoothing followed by Kauwahara filtering and local standard deviation approach for boundary detection. NIH FIJI/ImageJ tools are used to create the ground truth dataset. The results of this work indicate that detection of cell boundaries using segmentation approach even in the case of realistic measurement conditions is a challenging problem.
\end{abstract}


\section{Introduction}
Cell spreading and migration is a complex process in tumor development which occurs during cancer progression. Metastases are formed by cancer cells which evacuates from primary tumor. Those cells travel through circulatory system, attach to the vessel surface and then invade into intact tissues and cause growth of secondary tumor\cite{IEEEhowto:Hulkower2011}. Based on a growing body of evidences for last decades cancer research community raised opinion to develop the strategy concept for prevention of invading cancer cell to migrate and attach to target site in healthy tissue\cite{IEEEhowto:Coppola2008}. However, progress in this area is limited by necessity to collect data of large scale experiments and availability of few methods\cite{IEEEhowto:Valster2005}. One of such methods is direct study of cell spreading process on various types of 2D surface dedicated for cell culture. This in vitro experimental model is a good research tool and it reflects processes which play key role in metastasis progression in situ \cite{IEEEhowto:Dubin-Thaler2008}. In our previous results it was shown that there is a necessity of automated study tool implementation for analysis of cell spreading using full power of high-throughput microscopy. Combination of label-free techniques and automated image recognition is promising tool to study processes of cell spreading and motility\cite{IEEEhowto:Wollman2007}.

\section{Methods}
\subsection{Materials and data acquisition}
Live cell imaging was performed on 3T3 fibroblasts, U118 glyoblastoma, HT1080 fibrosarcoma and A549 non-small cell lung cancer cells maintained according to standard mammalian cell culture protocol. 3T3 cells are considered as normal cells, other cell lines origin from human tumors tissue. Cell lines were chosen because of their capability to attach and migrate on substrate. For microscopy observations, cells were detached from culturing flask and transferred to multi-well plate placed onto microscope stage. Data were collected as time-lapse microscopy image sequences using Carl Zeiss Axio Observer automated microscope equipped with heating incubating chamber and Hamamatsu ORCA V2 CMOS camera. Microscope was programmed to capture frames for every 10 minutes in DIC illumination mode, XY coordinates, focus and Kohler illumination was tuned before each record. There was a set of 4 regions of interest (ROI) each consisting from 2x2 fields of view (FOV) of 10x Objective. ROI was designated for selected treatment with microtubule inhibitors – nocodazole, taxol and vinorelbine (3 ROI) and for untreated control (1 ROI) to ensure absence of cell culturing artifacts. Cells for each record had common source (same passage and equal amount of suspension) and were plated immediately after detaching from growth flask. For each adjacent field of view there was up to 50 cells, cell clusters and debris. All images were acquired at CZI format (Carl Zeiss Image Data file type) using x10 objective. For comparison data obtained during this study were processed manually and semi qualitatively.

\subsection{Measurements}
The main method of analysis of spreading is based on description of cell area expansion using the measurements of cell perimeter alterations during recording. We can determine individual cell behavior pattern and population behavior pattern by analysis of the cell area changes. At the level of individual cell behavior of the spreading pattern vary widely. Thus relatively large cell samples (60 cells and more) are required to retrieval information about the entire cell population. There are three features of the cell behavior that are considered as most informative: area extension of cell, cell perimeter extension and circularity of a cell. Circularity is calculated via $circularity= 4\pi\frac{area}{perimeter^2}$\cite{IEEEhowto:Dubin-Thaler2008}, \cite{IEEEhowto:Aplin1983}.
This measure indicates perimeter a perfect circle at values of 1.0 and any decrease from this value reflect less circular shape.  This process indicates that cell is successfully attached to substrate, finished spreading, and start translocating on the substrate. During all these stages each cell maintains its shape by fine regulation of cytoskeleton proteins. This process may be easily corrupted, i.e. by microtubule inhibitors which is widely used in chemotherapy treatment \cite{IEEEhowto:fanale2015}. 
Next we extracted population behavior curves reflecting transition to completely spread cell: for each time point there was number of cells achieving 100\% area spreading. It was previously shown that normal 3T3 cell has high rates of both individual cell and that of population spreading, and further as expected cell were tole1rant for used 30 nM and 100 nM concentrations of drugss\cite{IEEEhowto:ASCB2016}%
Proposed method based on perimeter features is applicable for extraction data for other population properties: 1) cell cycle changes in entire population; 2) attachment prolongation period; and 3) elongation factor in population. In this case features collected on population level are more reliable compared to single cell because in situ interactions involves populations of cells and not only individual cells. In our previous study it was discussed about effect of spreading alteration on data retrieved from classic assay such as wound healing. It was noticed for A549 that taxol at low concentration causes an increase of spreading rates\cite{IEEEhowto:ASCB2016}. It was also reported by several authors that taxol may increase the speed of cell migration on the 2D substrate. However existing understanding on the information is fragmented and there is no unified opinion regarding this phenomena. Heterogeneous response on drug treatment when individual cells react on treatment and not in the same way as population may explain such phenomena. So far, it requires new tool development because of large amount of data to process which is one of the aims of this particular study.

\subsection{Cell segmentation at images}
Image sequence is set of time-lapse images of individual mammalian cell of different origin cultured on a 2D substrate. Prior to image processing all data was exported from CZI files to PNG sequences with ZEN lite software. Because of necessity to analyze individual cell behavior and presence of cell clusters interfering with recognition of individual cells, the sample sequence was cropped off manually with NIH FIJI/ImageJ image stack tools to a size of 300$\times$300 pixels. All units for measurement were converted to pixel units. Then original sequences of 16-bit PNG files where processed with Matlab. After initial step of normalization of image (gray scale was set between 0 and 1) and removal of shading artifacts, a flat field correction was performed on the images using the difference of Gaussians followed by gray scale normalization.

The first set of filtering operations is performed using the concept of G-neighbors\cite{IEEEhowto:G-neib1993}. It is based on the 3$\times$3 similarity mask where weight for the pixels given in the image is calculated using the similarity of the central pixel with that of the adjacent neighbors. To obtain an image pixel, all neighbors in 3$\times$3 array, which are smaller than the similarity threshold are multiplied with the filter weights and the obtained values are summed up to get the corresponding image pixel. The threshold is calculated as the square of the difference between the mean of maximum and mean of minimum pixel in the image. Next step includes Kuwahara neighborhood filter: 1) to smooth local contrast; 2) to remove any additional noise after G-neighbor smoothing. This was followed up with the local standard deviation performed on images as an option to reduce the impact of background pixels.

There were two options for processing the image further: (1) batch sample image processing to obtain preliminary data and (2) single sample image processing to perform segmentation adjustment. For batch processing there was direct transition to binarization after local standard deviation filter had been applied to the image. For single image processing we additionally performed bypass filtering of structures selecting too large and too low. At this stage we set filter properties manually to determine best options for segmentation. Threshold for binarization was also given manually. This stage was performed only for single cell image processing. Further segmentation was performed through morphological binary operations close> largest object filtration >fill holes> erode. Close and erode operation were performed using flat disk shape structuring elements with specified radius (in pixels). After this step there is additional largest element detection and accidentally appearing unconnected pixels were removed. Final step was measuring of the perimeter of a single object (cell).

 \begin{figure}[!htb]
   \centering
   \includegraphics[width=\linewidth]{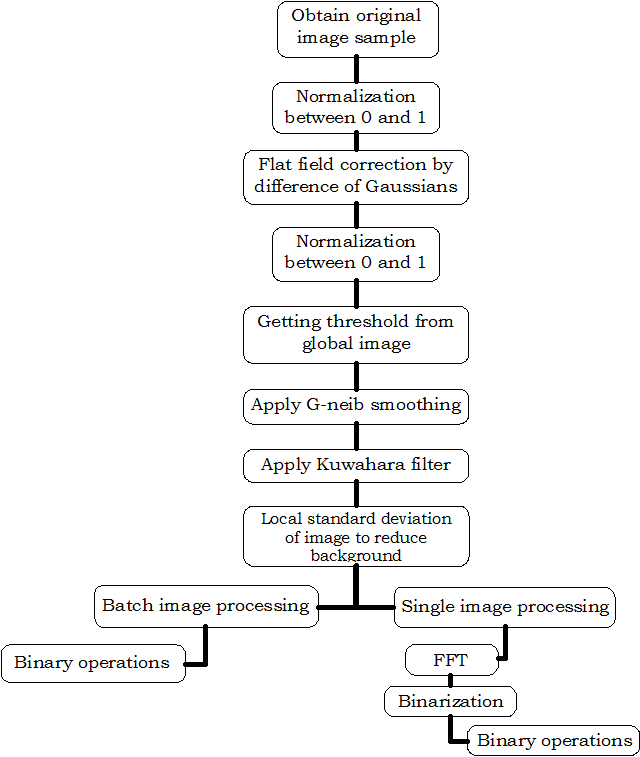} 
   \caption{Order of image processing operations. We applied processing in two ways: 1) initial batch processing with preset parameters of filtration. This part does not contained FFT filtration. 2) Individual image processing with addition of band-pass filter in the Furrier space (FFT).}
   \label{fig:algo}
  \end{figure}

\begin{figure}[h!]
\centering
\subfigure[{\tiny Frame 1, original image}]{\includegraphics[width=.07\textwidth]{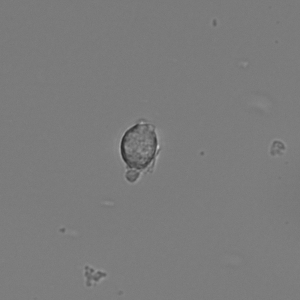}}
\subfigure[{\tiny Frame 2, original image}]{\includegraphics[width=.07\textwidth]{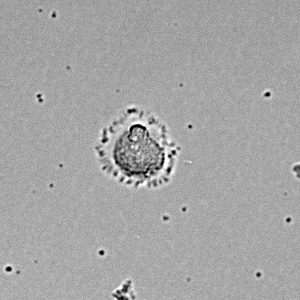}}
\subfigure[{\tiny Frame 9, original image}]{\includegraphics[width=.07\textwidth]{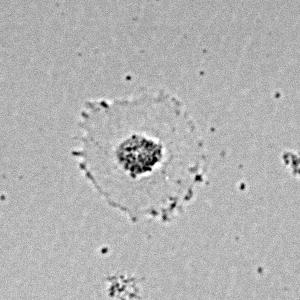}}
\subfigure[{\tiny Frame 15, original image}]{\includegraphics[width=.07\textwidth]{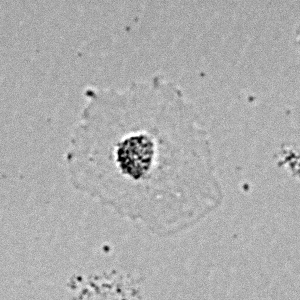}}
\subfigure[{\tiny Frame 25, original image}]{\includegraphics[width=.07\textwidth]{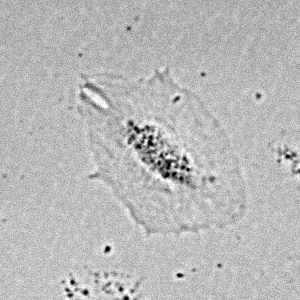}}
\subfigure[{\tiny Frame 40, original image}]{\includegraphics[width=.07\textwidth]{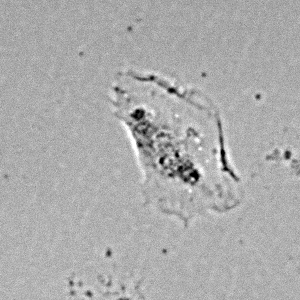}}
\subfigure[{\tiny Frame 1, algorithm segmentation}]{\includegraphics[width=.07\textwidth]{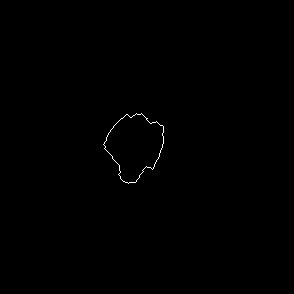}}
\subfigure[{\tiny Frame 2, algorithm segmentation}]{\includegraphics[width=.07\textwidth]{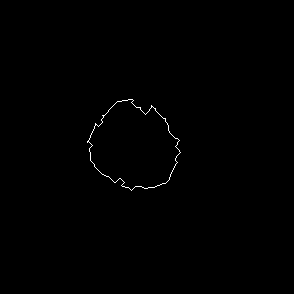}}
\subfigure[{\tiny Frame 9, algorithm segmentation}]{\includegraphics[width=.07\textwidth]{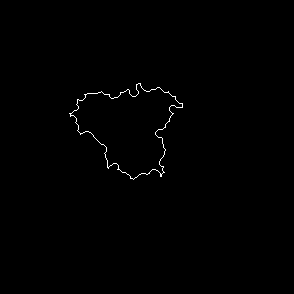}}
\subfigure[{\tiny Frame 15, algorithm segmentation}]{\includegraphics[width=.07\textwidth]{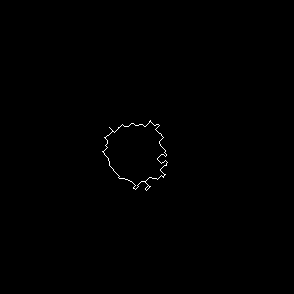}}
\subfigure[{\tiny Frame 25, algorithm segmentation}]{\includegraphics[width=.07\textwidth]{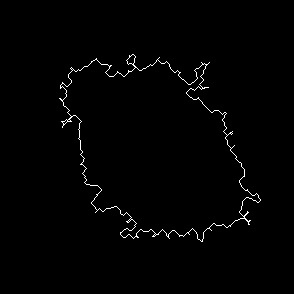}}
\subfigure[{\tiny Frame 40, algorithm segmentation}]{\includegraphics[width=.07\textwidth]{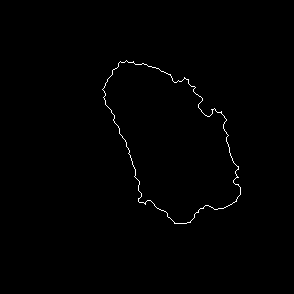}}
\subfigure[{\tiny Frame 1, manual segmentation}]{\includegraphics[width=.07\textwidth]{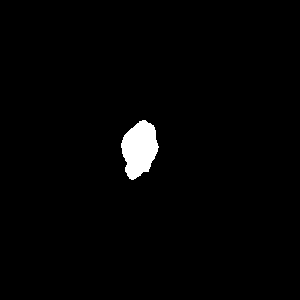}}
\subfigure[{\tiny Frame 2, manual segmentation}]{\includegraphics[width=.07\textwidth]{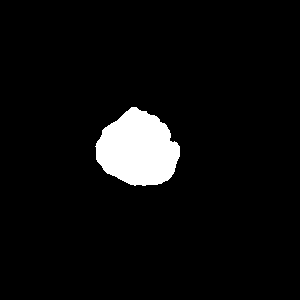}}
\subfigure[{\tiny Frame 9, manual segmentation}]{\includegraphics[width=.07\textwidth]{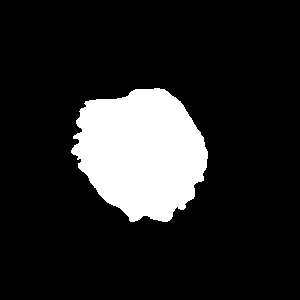}}
\subfigure[{\tiny Frame 15, manual segmentation}]{\includegraphics[width=.07\textwidth]{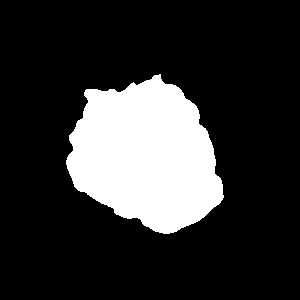}}
\subfigure[{\tiny Frame 25, manual segmentation}]{\includegraphics[width=.07\textwidth]{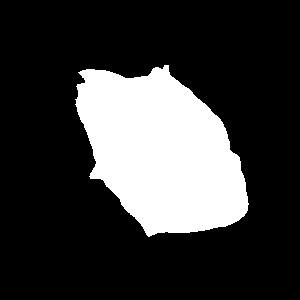}}
\subfigure[{\tiny Frame 40, manual segmentation}]{\includegraphics[width=.07\textwidth]{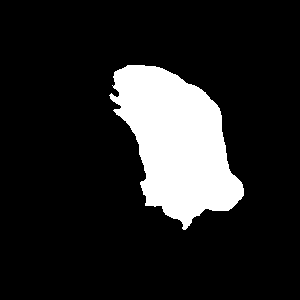}}
\caption{Original images sample sequence, results of algorithm processed and manually outlined cell contour. The image \textbf{i} and \textbf{j} of sequence show bad segmentation accuracy compared to manually outlined images, while rest of sequence has acceptable level.}
\label{fig:sequence}
\end{figure}

Original frames was obtained from U-118 cell line sample set. U-118 is human glyoblastoma cells widely used as model for drug testing. Also it has fibroblast like morphology and highly dynamic shape change pattern. From original microscopic images we created sets of frames containing single cell with manually outlined borders as ground truth dataset. This sets were used to have primary evaluation of segmentation successiveness. In result of described batch segmentation we obtained outline masks for each image in sequence with different accuracy of segmentation. Frames 1, 2, 9, 15, 20, 25, 40 of U-118 cell line sequence at Fig.\ref{fig:sequence} an example of drastic changes of segmentation accuracy compared to manual perimeter detection.

\begin{figure}[h!]
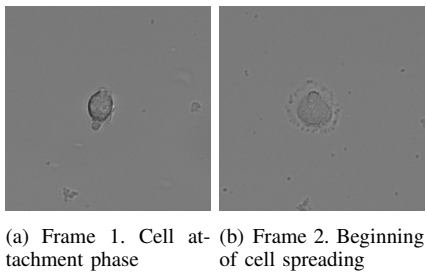

\centering
\subfigure[Frame 1. Cell attachment phase]{\includegraphics[width=.15\textwidth]{Fig1Fr1}}
\subfigure[Frame 2. Beginning of cell spreading]{\includegraphics[width=.15\textwidth]{Fig1Fr2}}
\caption{Original images of sequence is example of high dynamics of cell behavior affecting on contrast properties}\label{fig:orig}
\end{figure}

\begin{figure}[h!]
\centering
\subfigure[Flat field correction, Frame 1]{\includegraphics[width=.1\textwidth]{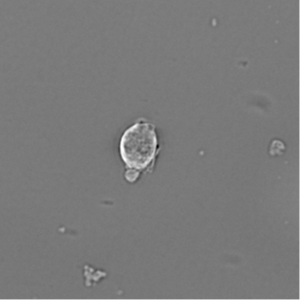}}
\subfigure[G-smothing, Frame 1]{\includegraphics[width=.1\textwidth]{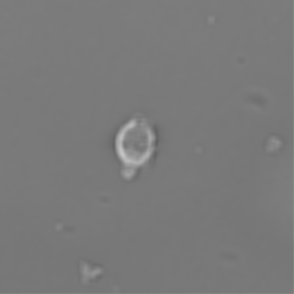}}
\subfigure[Kuwahara, Frame 1]{\includegraphics[width=.1\textwidth]{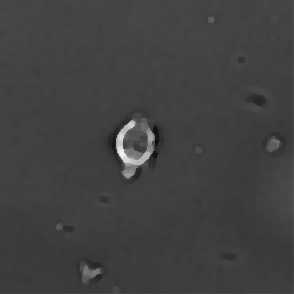}}
\subfigure[Reduce background, Frame 1]{\includegraphics[width=.1\textwidth]{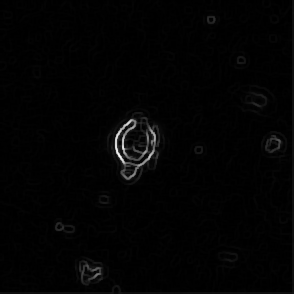}}
\subfigure[Flat field correction, Frame 2]{\includegraphics[width=.1\textwidth]{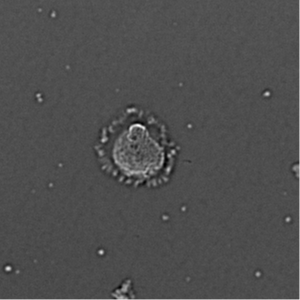}}
\subfigure[G-smothing, Frame 2]{\includegraphics[width=.1\textwidth]{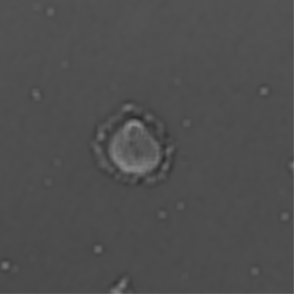}}
\subfigure[Kuwahara, Frame 2]{\includegraphics[width=.1\textwidth]{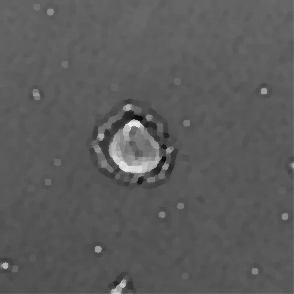}}
\subfigure[Reduce background, Frame 2]{\includegraphics[width=.1\textwidth]{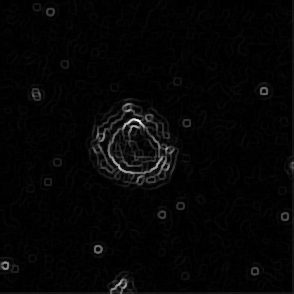}}
\caption{Result of filter applying on images of Frame 1 and 2 of sample sequence processing steps.}
\label{fig:frame2filterD}
\end{figure}

The Fig.\ref{fig:orig} is representing couple of frames of sample image sequence with relatively high contrast between foreground and can be segmented with almost 95\% accuracy with any method we used.
Frame 1 of sample sequence presented at Fig.\ref{fig:orig}(a) contain regions of hyper contrast pixels. In this case such condition does not affect on valid recognition since they are belong to object and present on margins of it and differs well from background. For Frame 2 there is absence of such pixels (Fig.\ref{fig:orig}(b)) and this conditions also makes segmentation possible in relatively inflexible filtering conditions. The Frame 2 has relatively low contrast however it is seen that after filtration applied on image we achieve good difference of core and cell body compared to background as it presented at Fig.\ref{fig:frame2filterD} . 

\begin{figure}[h!]
\centering
\subfigure[Threshold=0.01, Frame 1]{\includegraphics[width=.09\textwidth]{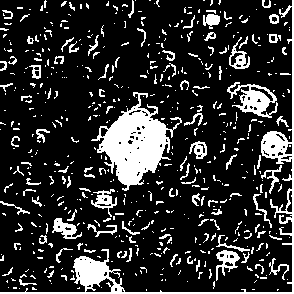}}
\subfigure[Result of segmentation, Frame 1]{\includegraphics[width=.09\textwidth]{Fig5F}}
\subfigure[Threshold=0.02, Frame 1]{\includegraphics[width=.09\textwidth]{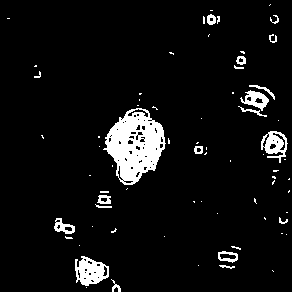}}
\subfigure[Result of segmentation, Frame 1]{\includegraphics[width=.09\textwidth]{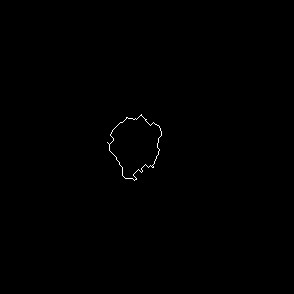}}
\subfigure[Manual drawing of cell contour, Frame 1]{\includegraphics[width=.09\textwidth]{Fig4A}}
\subfigure[Threshold=0.01, Frame 2]{\includegraphics[width=.09\textwidth]{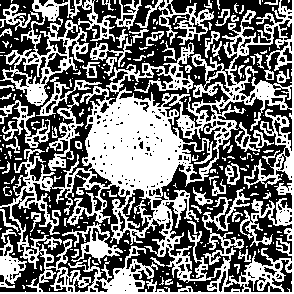}}
\subfigure[Result of segmentation, Frame 2]{\includegraphics[width=.09\textwidth]{Fig6F}}
\subfigure[Threshold=0.02, Frame 2]{\includegraphics[width=.09\textwidth]{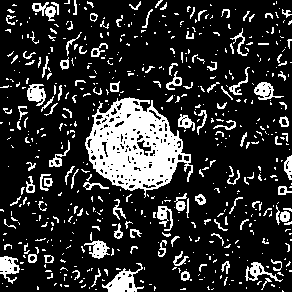}}
\subfigure[Result of segmentation, Frame 2]{\includegraphics[width=.09\textwidth]{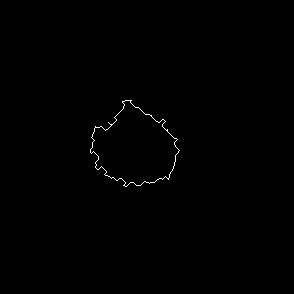}}
\subfigure[Manual drawing of cell contour, Frame 2]{\includegraphics[width=.09\textwidth]{Fig4B}}
\caption{Segmentation process and manual segmentation, of Frame 1 and 2 of sample sequence. Contours image(d) obtained after segmentation with different threshold values is close to manual segmentation(e) and can be considered as true detection}
\label{fig:segm2}
\end{figure}

For segmentation in this particular case we applied thresholds values 0.01 and 0.02, made after normalization and it has weak effect on final results Fig.\ref{fig:segm2}. Perimeter shape is different but still it is close to desired boundaries, detected by an eye. 

\begin{figure}[h!]
\centering
\subfigure[Flat field correction]{\includegraphics[width=.1\textwidth]{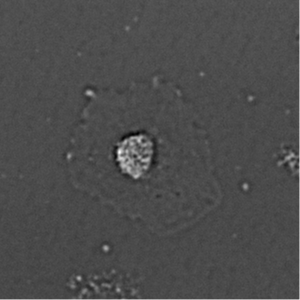}}
\subfigure[G-smothing]{\includegraphics[width=.1\textwidth]{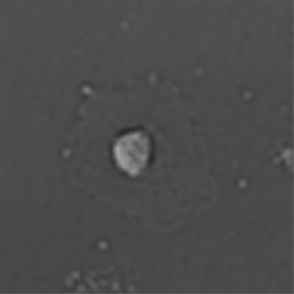}}
\subfigure[Kuwahara]{\includegraphics[width=.1\textwidth]{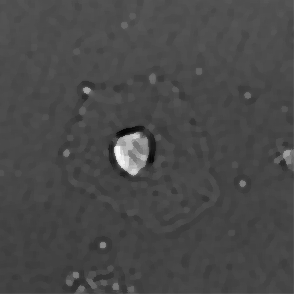}}
\subfigure[Reduce background]{\includegraphics[width=.1\textwidth]{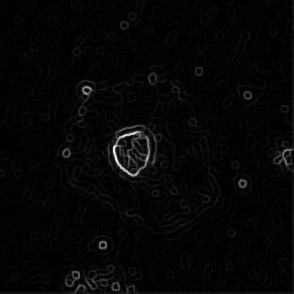}}
\caption{Contrast enhancement as result of filter applying on images of Frame 15.}
\label{fig:proc15}
\end{figure}

\begin{figure}[h!]
\centering
\subfigure[Threshold=0.01]{\includegraphics[width=.09\textwidth]{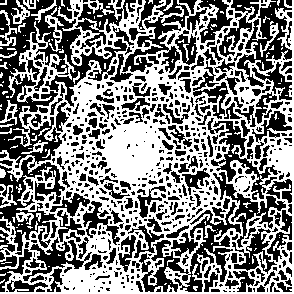}}
\subfigure[Result of segmentation]{\includegraphics[width=.09\textwidth]{Fig9F}}
\subfigure[Threshold=0.02]{\includegraphics[width=.09\textwidth]{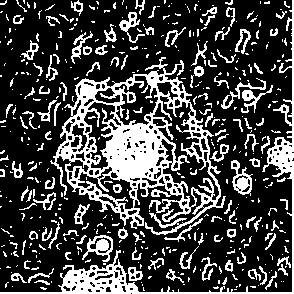}}
\subfigure[Result of segmentation]{\includegraphics[width=.09\textwidth]{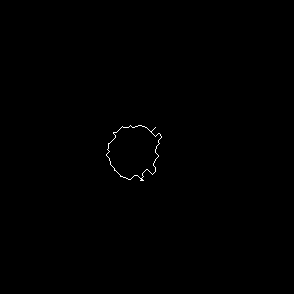}}
\subfigure[Manually defined perimeter]{\includegraphics[width=.09\textwidth]{Fig7M}}
\caption{Binarization of Frame 15 of sample sequence shows that there is boundaries available for segmentation. If compared images with values 0.01 and 0.02 of threshold show that proper value is between them, closer to 0.02 rather than 0.01}
\label{fig:segm15}
\end{figure}

As it was noticed  Fig.\ref{fig:sequence}(d) is example of low accuracy segmentation. The highly contrast cell core on this image attracts attention, while cell body and boundaries remains shattered. This contrast difference caused by cell morphology since cell itself has no more lens shape and changes to sombrero-hat shape affecting on contrast distribution.
Fig.\ref{fig:proc15} demonstrate changes of cell image during processing. As it seen there is high contrast core(actual cell nuclei zone) and almost invisible by eye body of cell with contrast same as in background when it seen by eye. After filtration with G-neighbor and Kuwahara it is appear to became more contrast. However because of  values for binarization applied globally for rest of images in sequence it produce false negative recognition \ref{fig:segm15}. 

\begin{figure}[h!]
\centering
\subfigure[Flat field correction]{\includegraphics[width=.1\textwidth]{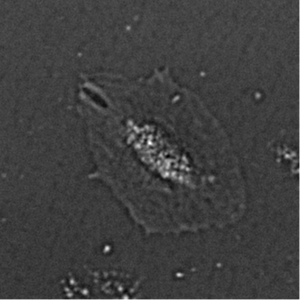}}
\subfigure[G-smothing]{\includegraphics[width=.1\textwidth]{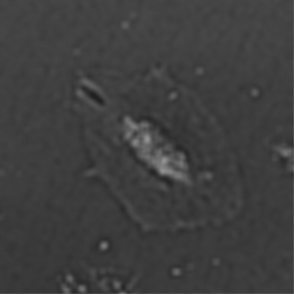}}
\subfigure[Kuwahara]{\includegraphics[width=.1\textwidth]{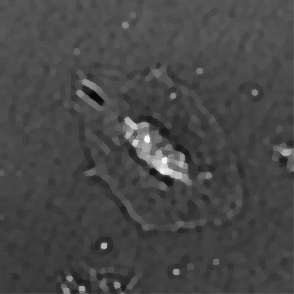}}
\subfigure[Reduce background]{\includegraphics[width=.1\textwidth]{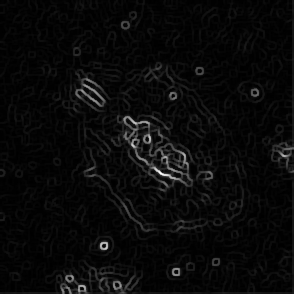}}
\caption{Filtration process, frame 25.}
\label{fig:proc25}
\end{figure}

\begin{figure}[h!]
\centering
\subfigure[Threshold=0.01]{\includegraphics[width=.09\textwidth]{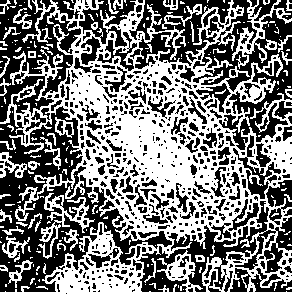}}
\subfigure[Result of segmentation]{\includegraphics[width=.09\textwidth]{Fig12F}}
\subfigure[Threshold=0.02]{\includegraphics[width=.09\textwidth]{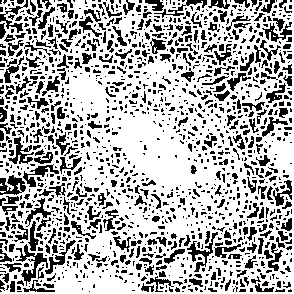}}
\subfigure[Result of segmentation]{\includegraphics[width=.09\textwidth]{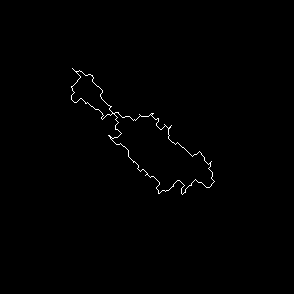}}
\subfigure[Manually defined perimeter]{\includegraphics[width=.09\textwidth]{Fig10M}}
\caption{Binarization of Frame 25 of sample sequence shows true or false of boundaries detection case. This difference depend on threshold values where we obtain good segmentation at 0.01 value rather than 0.02. Meanwhile value providing more accurate segmentation is >0.01}
\label{fig:segm25}
\end{figure}

Next case is Frame 25 of sample sequence comparison of between values 0.01 and 0.02 of threshold show that proper value is around 0.01. Fig.\ref{fig:proc25} shows that itself cell have homogeneous contrast. With general binarization it is lead to relatively good accuracy for 0.01 threshold value or leading to false negative segmentation at 0.02 threshold value as it resent at Fig.\ref{fig:segm25}.

Images in sequence was inspected further and false recognition is appear on images with contrast distribution artifacts described above. Artifacts includes heterogeneous contrast distribution in cells, caused by shape or debris stuck on cell or appearing in field of view for several frames. Thus we applied low-pass and high-pass filtration and individual thresholds to improve segmentation by removing unnecessary structures. 

\section{Results}
Original images were preliminary processed as batch sequence with code containing G-neighbor and Kuwahara filtration only or individual sample processing with adjustable bypass FFT filtering with preset global values. Global threshold for binarization were determined manually and was in range of values between 0.01-0.02 for different sets. Binary operations sequence for segmentation was performed next way: erode>close>fill holes>erode with structuring elements size up to 3 pixel. To remove accidental particles there was largest object filtering performed and it’s perimeter was extracted. In result we obtained outline masks for each image in sequence with different accuracy of segmentation. 

\begin{figure}[h!]
\centering
\includegraphics[width=\linewidth]{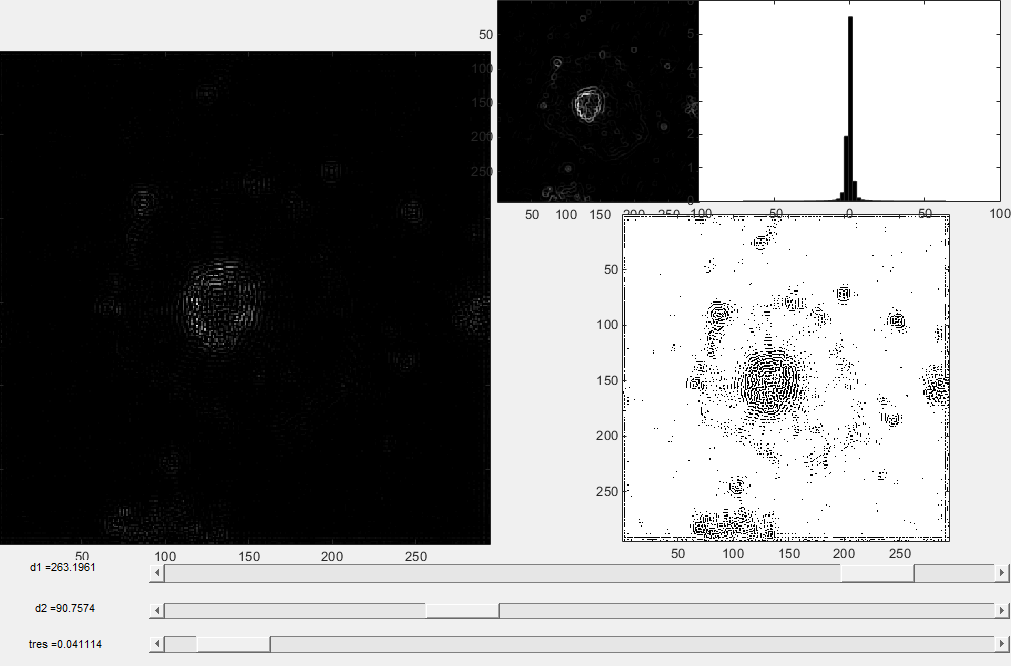}
\caption{Dialog window for FFT filtering and conversion to binary image. Where are D1 and D2 represent the size of filter. Maximum and minimum values is depend on image window. Third option for choice is Threshold value.}
\label{fig:slider}
\end{figure}

\begin{figure}[h!]
\centering
\subfigure[Filter settings: D1=280.1423; D2=0.74; threshold=0.04]{\includegraphics[width=.1\textwidth]{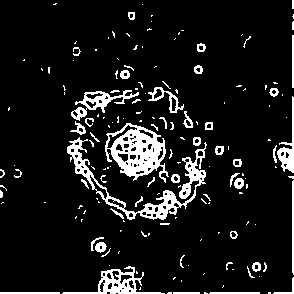}}
\subfigure[Filter settings: D1=65.5; D2=0.8688; threshold=0.0305]{\includegraphics[width=.1\textwidth]{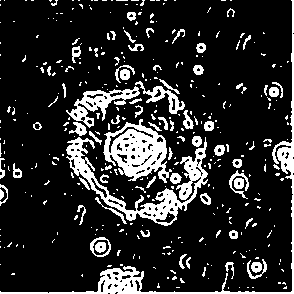}}
\subfigure[Filter settings: D1=263.1961; D2=90,7574; threshold=0.0417]{\includegraphics[width=.1\textwidth]{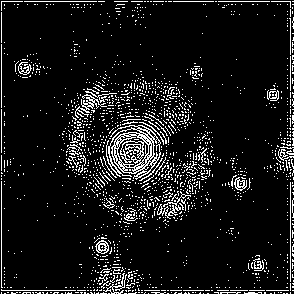}}
\subfigure[Original unprocessed image]{\includegraphics[width=.1\textwidth]{U-118-1_0009}}
\subfigure[Filter option]{\includegraphics[width=.1\textwidth]{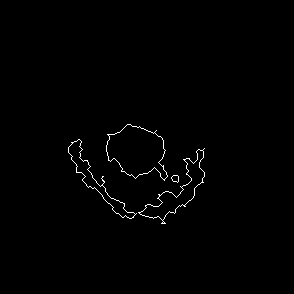}}
\subfigure[Filter option]{\includegraphics[width=.1\textwidth]{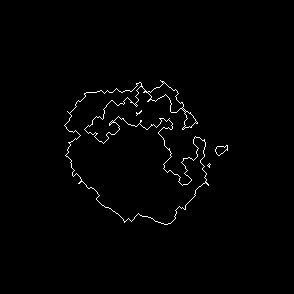}}
\subfigure[Filter option]{\includegraphics[width=.1\textwidth]{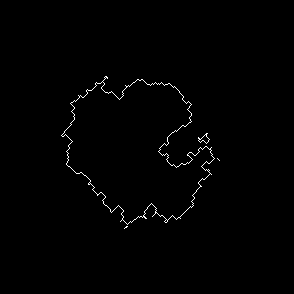}}
\subfigure[Overlay of manually defined perimeter]{\includegraphics[width=.1\textwidth]{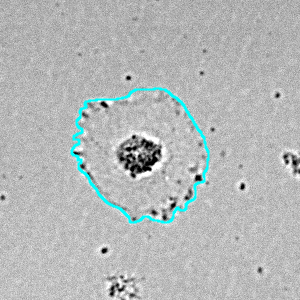}}
\caption{Filtration process of Frame 9 with manually chosen FFT filter properties(a-c). Results of segmentation (e-g) with manually outlined image(h).}
\label{fig:result}
\end{figure}

Perimeter values measured with batch sequence processing (without FFT) is results of inflexible filter parameter. Each image may be considered as model object with: background which must be filtered out, cell itself to be segmented and measured. Most people usually recognize cell body surrounded by thin black collar or aureole related to cell lamella (thin sheet of cell membrane protrusions not always recognized by novice user). It may be physically absent due to cell rounding process (Frame 1), when cell detach from substrate. And cell core is refers to cell nuclei zone however it is contain not only nuclei. The core is central part of cell image, it is thicker than lamella which causes contrast difference. If simplify cell shape maybe described as positive meniscus lens (flat part is attached to substrate) with zones of different contrast (caused by presence of cellular structures). Uniformed binary operations cannot provide same level of detection accuracy. Thus proper binarization procedure and filtering operations prior to binarization were determined as important part affecting on segmentation accuracy. However further we find out that binarization procedures cannot be standardized and must be set up for each image. 
During inspection of image properties false recognition was noticed for images with higher difference of pixel values compared to control. Adjustable FFT (with real time results exhibition) was performed on such images to extract object from background (Fig.\ref{fig:slider}). Filtration with different options of FFT applied on sample image with bad detection accuracy (Frame 9) represented at Fig.\ref{fig:result}(a-c) show that we can find parameters of filtration providing good level of segmentation. Threshold was adjusted manually for each image. 

Image (f) of Fig.\ref{fig:result} shows most similarity to manually detected cell contour. But it has not connected contour which is considered as false detection. However compared to results after batch processing it has higher accuracy rate.

\section{Conclusion}
According to this result image processing of such sequences may be performed with adaptive filtering and binarization which can give high level of accuracy for all images in sequence. Ideal case of sample sequence of cell images contains homogeneous background without outline structures(debris) or illumination imperfections and cell itself, present by body(limited by lamella) and core(nuclei and surrounding compartments) where weights of body+core <not equal> to background for each image. If pixel weight of core > body+background there is false detection but if (core=body) or (core+body) not equal to background there is true detection. In opposite real samples processing is challenging due of presence of images with varying contrast features produced by object itself (contrast distribution, illumination or cell culturing artifacts, etc.). Here is two limitation factor where binarization appear to be less complex part of segmentation. However conventional approaches such as Otsu method does not lead to success in this particular case. As it was described for Frame 25 itself it may produce false recognition while filtration steps produce good results. There are several approaches exist for text recognition binarization utilizing pixel by pixel or region by region approach\cite{IEEEhowto:Sauvola2000}. This part remains disputable since cell images has lower contrast compared to text samples. It was shown at Fig.\ref{fig:result} individually applied filtration can in particular solve the problem. However user still need to understand process of further image generation which lead to complicated manipulation to extract body from substrate. Method we proposed is combination of several existing methods to support keystone concept of our study based on simulation of the processes of object boundary recognition performed by human visual analyzer. Collected samples of images shows that cell borders presented as various patterns of shades and light gradient distributions, also it may contain different spatial structures changing in time domain. Understanding of a nature of the true object in images requires strong background in life sciences, however recognition of object boundaries themselves in this images does not requires special knowledge. Object recognition is made by human utilize cascade of non-linear filtering operations applied to the primary image by visual analyzer running in visual cortex of human brain\cite{IEEEhowto:DiCarlo2012}. In this work we made attempt to perform reduced set of operations with non-linear filtering but with assistance of neighborhood operation and high and low pass filtering to achieve cell border segmentation in a way close to that performed in brain. Obtained results are still doubtful and we have to speculate about filtering method selection further. However it is evident that successful image segmentation with boundary detection may be achieved in certain conditions.

\section{Acknowledgements}
We thank to Olga Krestinskaya, Master student of Department of Electrical Engineering, School of Engineering of Nazarbayev University for batch image processing code writing and to Arshat Urazbaev researcher of National Laboratory Astana, for assistance in G-neighbor and real time FFT filtering code writing.


\begin{thebibliography}{1}
\bibitem{IEEEhowto:Hulkower2011}
K.~Hulkower, R.~Herber, \emph{Cell migration and invasion assays as tools for drug discovery}, Pharmaceutics 2011, 3(1): 107-124; doi: 10.3390/pharmaceutics3010107
\bibitem{IEEEhowto:Coppola2008}
J.~Coppola, M.~Bhojani, B.~Ross, A.~Rehemtulla, \emph{A small-molecule furin inhibitor inhibits cancer cell motility and invasiveness}, Neoplasia 2008, 10(4): 363-370; doi: 10.1593/neo.08166
\bibitem{IEEEhowto:Valster2005}
A.~Valster, N.~Tran, M.~Nakada, M.~Berens, A.~Chan, M.~Symons,  \emph{Cell migration and invasion assays}, Methods 2005, 35: 208-215; doi: 10.1016/j.ymeth.2005.08.001
\bibitem{IEEEhowto:Dubin-Thaler2008}
B.~Dubin-Thaler, J.~Hofman, Y.~Cai, H.~Xenias, I.~Spielman, A.~Shneidman, L.~David, H.~Döbereiner, C.~Wiggins, M.~Sheetz, \emph{Quantification of cell edge velocities and traction forces reveals distinct motility modules during cell spreading}, PLoS ONE 2008, 3(11):e3735; doi:10.1371/journal.pone.0003735.
\bibitem{IEEEhowto:Wollman2007}
R.~Wollman and N.~Stuurman, \emph{High throughput microscopy: from raw images to discoveries}, Journal of Cell Science 2007, 120: 3715-3722; doi: 10.1242/jcs.013623.
\bibitem{IEEEhowto:Dubin-Thaler2004}
B.~Dubin-Thaler , G.~Giannone, H-G.~Döbereiner, M.~Sheetz, \emph{Nanometer Analysis of Cell Spreading on Matrix-Coated Surfaces Reveals Two Distinct Cell States and STEPs}, Biophys J. 2004 Mar; 86(3): 1794–1806; doi:  10.1016/S0006-3495(04)74246-0.
\bibitem{IEEEhowto:Aplin1983}
J.~Aplin, W.~Bardsley, V.~Niven, \emph{Kinetic analysis of cell spreading. II. Substratum adhesion requirements of amniotic epithelial (FL) cells}, J Cell Sci. 1983 May;61:375-388;
\bibitem{IEEEhowto:fanale2015}
D.~Fanale, G.~Bronte, F.~Passiglia, V.~Calò, M.~Castiglia, F.~Di Piazza, N.~Barraco, A.~Cangemi, MT.~Catarella, L.~Insalaco, A.~Listì, R.~Maragliano, D.~Massihnia, A.~Perez, F.~Toia, G.~Cicero, V.~Bazan, \emph{Stabilizing versus destabilizing the microtubules: A double-edge sword for an effective cancer treatment option?}, Analytical Cellular Pathology 2015; doi:  10.1155/2015/690916.
\bibitem{IEEEhowto:ASCB2016}
S.~Kauanova, A.~Tvorogova, A.~Kakpenova, A.~Balabiyev, M.~Abdukassimova, S.~Mussakhan, I.~Vorobjev, \emph{Spreading and motility of normal and cancer cells under the action of anti-microtubule drugs – dose-response relation}, Mol Biol Cell 2016;27(25):4606 (abstract P1087); doi:10.1091/mbc.E16-10-0736.
\bibitem{IEEEhowto:G-neib1993}
T.~Boult, E.~Terrance, R.~Melter, F.~Skorina  and I.~Stojmenovic, \emph{G-neighbors}, Proc. SPIE 2060, Vision Geometry II, 96 (December 1, 1993); doi:10.1117/12.165007.
\bibitem{IEEEhowto:DiCarlo2012}
J.~DiCarlo, D.~Zoccolan, N.~Rust, \emph{"How does the brain solve visual object recognition?"}, Neuron, 2012 February 9; 73(3): 415-434. doi:10.1016/j.neuron.2012.01.010.
\bibitem{IEEEhowto:Sauvola2000}
J.~Sauvola, M.~Pietikäinen, \emph{Adaptive document image binarization}, Pattern Recognition, 2000;33(2):225-236. doi:10.1016/S0031-3203(99)00055-2.
\end{thebibliography}
\end{document}